%% file: MICCAI2026-main_conference_paper_template.tex
\documentclass[runningheads]{llncs}
\usepackage[T1]{fontenc}
\usepackage{graphicx,verbatim}
\usepackage{amsmath}
\usepackage{algorithm}
\usepackage{algpseudocode}
\usepackage{amsmath}
\usepackage{amssymb}
\usepackage{booktabs}
\usepackage{booktabs}
\usepackage{orcidlink}
\usepackage{multirow}
\usepackage{graphicx}
\usepackage{xcolor}
\usepackage{amsmath,amssymb}
\begin{document}
\title{HPOQuest: A Rare-Disease Diagnostic Agent Using Active Phenotype Acquisition}
\titlerunning{HPOQuest}
\author{
Kamilia Zaripova\inst{1,2,3,4}\orcidlink{0000-0002-8200-0818} \and 
Nassir Navab\inst{1,3}\orcidlink{0000-0002-6032-5611} \and \\
Azade Farshad\inst{1,3,5}$^{\dagger}$\orcidlink{0000-0002-1080-1587} \and
Annalisa Marsico\inst{2}$^{\dagger}$\orcidlink{0000-0003-0327-5555}
}
\authorrunning{K. Zaripova et al.}
\institute{
Technical University of Munich, Munich, Germany \and
Computational Health Center, Helmholtz Center Munich, Oberschleißheim, Germany \and
Munich Center for Machine Learning (MCML), Munich, Germany \and
Munich Data Science Institute (MDSI), Munich, Germany \and
Aalto University, Espoo, Finland \\
\email{kamilia.zaripova@tum.de}
}
{\renewcommand{\thefootnote}{}\footnotetext{$^{\dagger}$ Shared last authorship}}
  
\maketitle              
\begin{abstract}
More than 300 million people worldwide are affected by one of over 7,000 known
rare diseases, yet diagnosis remains difficult because patients initially
present with incomplete and heterogeneous phenotypes. We present
\textbf{HPOQuest}, a training-free framework for sequential phenotype
acquisition in rare-disease diagnosis. Starting from a small set of observed patient phenotypes, HPOQuest maintains a
probabilistic disease ranking and iteratively selects informative follow-up
questions to support clinicians during patient assessment. Confirmed phenotypes update the disease ranking, while all responses update the candidate question set. Across four benchmark cohorts, HPOQuest
substantially improves diagnosis from sparse initial phenotypes, with gains of
up to $30\%$  points at Recall@1 and $45\%$  points at
Recall@5. These results demonstrate that sequential
phenotype acquisition can substantially improve rare-disease diagnosis from
limited initial clinical evidence.

\keywords{Rare disease diagnosis, Sequential phenotype acquisition, Human Phenotype Ontology, Active information acquisition}

\end{abstract}

\input{chapters/intro}
\input{chapters/method}

\input{chapters/experemental_setup}
\input{chapters/results_dis}

\input{chapters/conclusion}

\begin{credits}
\subsubsection{\discintname}
The authors have no competing interests to declare that are relevant to the content of this article.
\end{credits}
\bibliographystyle{splncs04}
\bibliography{ref}
\end{document}

%% file: chapters/intro.tex
\section{Introduction}

Although individually rare, each affecting fewer than 1 in 2,000 people in
Europe, rare diseases collectively affect more than 300 million people
worldwide and remain difficult to diagnose. Clinicians typically begin with a
few signs and symptoms and refine the differential through targeted questions,
examinations, and tests. Patient findings are commonly represented using the
Human Phenotype Ontology (HPO), a hierarchy of more than 20,000 phenotypic
abnormalities. However, extensive overlap among diseases and variation among
patients with the same disease leave many diagnoses plausible from only a few
initial terms. These terms also vary in diagnostic value: specific findings may
identify the relevant disease family, whereas common, nonspecific, or incidental
findings may provide little guidance. 

Current rare-disease systems primarily diagnose from observed phenotype
profiles or augment diagnosis with retrieval and LLM reasoning, but do not
explicitly address sequential, response-dependent phenotype acquisition. Existing rare-disease methods primarily model disease--phenotype associations
or generate diagnoses assuming fully supplied phenotype profiles
\cite{chen2026phenoss,zhao2026agentic}. RareBench~\cite{chen2024rarebench}
benchmarks phenotype extraction and rare-disease diagnosis, while systems such
as DeepRare~\cite{zhao2026agentic} and
RDguru~\cite{yang2024rdguru} combine LLMs with phenotype processing,
retrieval, and diagnostic tools. DeepRare diagnoses from a supplied phenotype profile, whereas RDguru performs
interactive phenotype acquisition but starts from 50\% of the recorded
phenotypes and relies on a multi-component GPT-4-based agent.
PhenoDP~\cite{wen2025phenodp} recommends one additional phenotype from partially observed profiles without
modeling sequential patient-dependent questioning.
\\
\indent Clinical questioning has been studied by MediQ and MedKGI through free-form
question asking and information-guided test selection~\cite{li2024mediq,wang2025medkgi}. MedClarify~\cite{wong2026medclarify}
similarly selects free-form questions using expected information gain.
LA-CDM~\cite{bani2025language} learns cost-efficient test selection using
reinforcement learning within four abdominal diseases, while other systems
learn diagnostic trajectories or employ specialized diagnostic
agents~\cite{hsu2026medaction,sanghvi2026medxagent}. Earlier active feature
acquisition methods rely on model attribution~\cite{vivar2020peri} or
information-theoretic selection~\cite{he2022bsoda}, whereas
ASIG~\cite{hartmann2026amortising} learns an LLM-based acquisition policy on
20 Questions and transfers it to MediQ. Together, these methods address related information-acquisition problems but
not sequential, response-dependent phenotype selection over the HPO hierarchy
across thousands of rare diseases.
\\
\indent Selecting the next phenotype remains challenging because the HPO contains more
than 20,000 hierarchically related terms, diseases share overlapping phenotype
profiles, and each response changes both the candidate diseases and the valid
follow-up questions. HPOQuest addresses this setting by starting
from the same three seed phenotypes for every patient, compared with median complete-profile sizes of 9--17.5 HPO terms. It iteratively selects a
limited number of valid HPO questions, and updates both the candidate diseases
and the available follow-up questions after each response. Its training-free
acquisition policy prioritizes questions using the current candidate diseases,
disease--phenotype association frequencies, and phenotype specificity.
\\
\indent HPOQuest uses a locally deployable, in-clinic, open-weight model and compares
all question selection methods using the same inputs, budget, responses,
retrieval, and diagnostic components. We contribute
(i) an ontology-constrained framework for phenotype acquisition from
retrospective HPO records, where three-way responses expand the candidate
question set toward confirmed descendants or prune absent subtrees;
(ii) a training-free, candidate-anchored policy for hierarchical phenotype
acquisition; and (iii) a controlled benchmark demonstrating when active phenotype acquisition
improves rare-disease diagnosis. When sparse no-acquisition diagnosis is already informative, HPOQuest remains
competitive with the strongest baseline across all evaluated cutoffs. When both
sparse baselines fail, as on the MME cohort, HPOQuest improves recall by
$30.0$--$45.0$ percentage points. HPOQuest therefore complements accessible
LLM and retrieved knowledge while providing its largest gains when that
knowledge is insufficient.

%% file: chapters/method.tex
\providecommand{\HPO}{\ensuremath{\mathcal{H}}}
\providecommand{\DIS}{\ensuremath{\mathcal{D}}}
\providecommand{\CONF}{\ensuremath{\mathcal{C}}}
\providecommand{\DEN}{\ensuremath{\mathcal{N}}}
\providecommand{\SEED}{\ensuremath{\mathcal{S}}}
\providecommand{\pbg}{\ensuremath{\hat{p}}}

\begin{figure*}[t]
\centering
\includegraphics[width=1\textwidth]{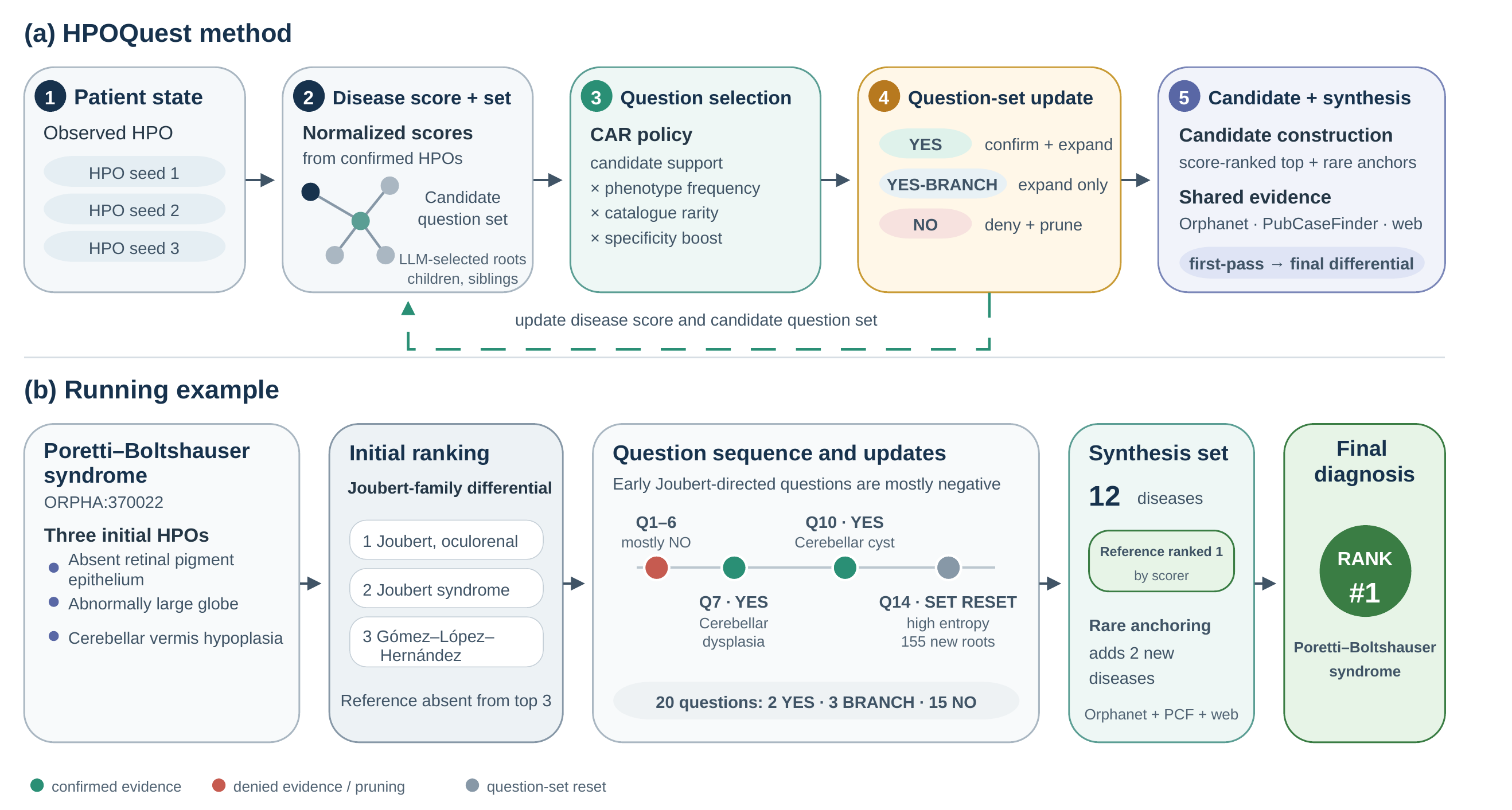}
\caption{
HPOQuest. (a) Starting from three seed HPO phenotypes, HPOQuest iteratively updates disease scores and the ontology-constrained candidate question set. The candidate-anchored rarity (CAR) policy selects follow-up questions, responses update both the disease ranking and candidate set, and the acquired phenotypes are combined with shared biomedical evidence to produce the final ranked differential diagnosis. (b) Representative diagnostic trajectory for a patient with Poretti--Boltshauser syndrome, showing how sequential phenotype acquisition eliminates competing diagnoses and promotes the reference disease to rank 1. }
\label{fig:hpoquest}
\end{figure*}

\section{Method}
\label{sec:method}
The Human Phenotype Ontology (HPO) is a directed acyclic graph whose nodes
represent clinical abnormalities and whose edges connect general phenotypes to
more specific ones. HPOQuest uses these terms as candidate questions. Starting
from a small set of observed phenotypes, it repeatedly ranks diseases,
selects valid follow-up questions, and updates both the disease ranking and
candidate question set after each response. The acquired phenotypes are then
combined with retrieved biomedical evidence to produce a ranked differential
diagnosis.

\vspace{2pt} \noindent { \textit {\textbf{Patient state and disease posterior.}}
HPOQuest begins with only $K$ observed seed phenotypes and iteratively
acquires additional evidence. Confirmed phenotypes are used to rank diseases,
and the ranking is updated after every new confirmation to guide question
selection. Let $\HPO$ denote the set of HPO terms and $\DIS$ the disease
catalogue. Each disease $d\in\DIS$ is associated with an HPO profile
$\operatorname{profile}(d)$ and phenotype frequencies $P(h\mid d)$. For
patient $p$, $\HPO^+(p)$ denotes the complete recorded positive phenotype set,
with $\SEED\subseteq\HPO^+(p)$ the initially observed seed phenotypes. HPOQuest
may ask at most $B$ additional phenotype questions.
During evaluation, responses are generated deterministically from the recorded
patient profile:

\begin{equation}
\label{eq:oracle}
a(h)=
\begin{cases}
\text{\textsc{yes}},
& h\in\HPO^+(p),\\
\text{\textsc{yes-branch}},
& h\notin\HPO^+(p)
  \ \text{and}\
  \operatorname{Desc}(h)\cap\HPO^+(p)\neq\emptyset,\\
\text{\textsc{no}},
& \text{otherwise}.
\end{cases}
\end{equation}
Because the response simulator is defined only with respect to the recorded positive phenotype set $\HPO^+(p)$, it adopts a closed-world assumption: terms absent from $\HPO^+(p)$, with no recorded descendant, are treated as negative. Here, $\operatorname{Desc}(h)$ contains the more specific HPO terms below $h$. A \textsc{yes} confirms $h$; \textsc{yes-branch} indicates a recorded descendant; and \textsc{no} denies $h$. We maintain confirmed and denied sets $\CONF$ and $\DEN$, initialized as $\CONF=\SEED$ and $\DEN=\emptyset$; a \textsc{yes-branch} term is added to neither set.

Confirmed phenotypes contribute weighted evidence for associated diseases. The
weight $w_h$ downweights generic seed phenotypes, while phenotypes confirmed
during questioning always receive unit weight:

We first measure how widely phenotype $h$ occurs across disease profiles:
\begin{equation}
\pbg(h)
=
\frac{
\left|\{d\in\DIS:h\in\operatorname{profile}(d)\}\right|
}{
|\DIS|
},
\qquad
w_h=
\begin{cases}
\gamma,
& h\in\SEED\ \text{and}\ \pbg(h)>\theta_s,\\
1,
& \text{otherwise}.
\end{cases}
\end{equation}

Here, $\pbg(h)$ is the fraction of diseases associated with $h$, $\theta_s$ identifies generic seed phenotypes, and $\gamma\in(0,1)$ is their downweighting factor. If all seeds exceed $\theta_s$, the least common seed retains unit weight. Phenotypes confirmed during questioning always receive unit weight. The phenotype
frequencies are combined into a weighted log-compatibility score
$\ell(d\mid\CONF)$, where a higher value indicates that disease is more
consistent with the currently confirmed phenotypes:

\begin{equation}
\ell(d\mid\CONF)
=
\sum_{h\in\CONF}w_h\log P(h\mid d),
\qquad
\pi(d\mid\CONF)
=
\frac{\exp\!\left(\ell(d\mid\CONF)\right)}
{\sum_{d'\in\DIS}\exp\!\left(\ell(d'\mid\CONF)\right)}.
\end{equation}

The softmax normalization converts these compatibility scores into a
distribution $\pi(d\mid\CONF)$ over the disease catalogue, which is used to rank diseases and guide
question selection. Only confirmed phenotypes modify this score; denied
phenotypes are used for pruning questions and during final diagnostic
synthesis.

\vspace{2pt} \noindent\vspace{2pt} \noindent { \textit {\textbf{Ontology-constrained question selection.}}}
\label{sec:acquisition}
At step $t$, the candidate question set $\mathcal{F}_t\subset\HPO$ contains
the HPO terms that may be queried next. It is initialized with the children
and siblings of the seed terms. To cover relevant regions beyond these local
branches, one LLM call selects additional top-level HPO categories from a
fixed list, excluding categories that already contain a seed. The LLM cannot
generate new HPO terms or select subsequent questions.

Before questioning, we issue one web search for each of the $K$ seed
phenotypes. Question selection itself uses only the HPO structure and the
disease--phenotype associations and frequencies.
Let $\mathcal{T}_M$ be the $M$ highest-ranked diseases under
$\pi(d\mid\CONF)$. We quantify the catalogue rarity of phenotype $h$ by

\begin{equation}
r(h)=-\log\max\{\pbg(h),\phi\},
\end{equation}

and identify associations that are frequent within a disease but uncommon
across the catalogue:

\begin{equation}
z(h,d)=
\begin{cases}
1,
& P(h\mid d)\geq\theta_f
  \ \text{and}\ \pbg(h)\leq\theta_r,\\
0,
& \text{otherwise}.
\end{cases}
\end{equation}

The acquisition score combines the current disease ranking,
within-disease phenotype frequency, catalogue rarity, and a specificity boost
for phenotypes that are common within a disease but rare overall. The
hyperparameters $\lambda\ge0$ and $\beta\ge0$ control the influence of the
disease ranking and specificity boost. Our candidate-anchored rarity (CAR)
policy assigns each eligible question $h\in\mathcal{F}_t$ the score

\begin{equation}
\label{eq:score}
s(h)
=
\sum_{d\in\mathcal{T}_M}
\underbrace{\left(1+\lambda\pi(d\mid\CONF)\right)}_{\text{current disease rank}}
\,
\underbrace{P(h\mid d)}_{\text{within-disease frequency}}
\,
\underbrace{r(h)}_{\text{catalogue rarity}}
\,
\underbrace{\left(1+\beta z(h,d)\right)}_{\text{specificity boost}}.
\end{equation}

Higher scores identify phenotypes that are both characteristic of the current
candidate diseases and discriminative across the disease catalogue. Questions
with $s(h)\ge\tau$ are selected in descending score order, up to
$\min\{b,B-q\}$, where $b$ is the per-round limit and $q$ the number of
questions already issued.


\vspace{2pt} \noindent { \textit{\textbf{Question-set update and stopping.}}}
\label{sec:frontier-update}
A \textsc{yes} or \textsc{yes-branch} response adds the queried term's
children to $\mathcal{F}_t$. A \textsc{no} response removes the queried term
and all its descendants. Queried and removed terms cannot be selected again.

Let $\mathcal{T}_L$ contain the $L$ highest-ranked diseases and let
$\bar{\pi}(d)=\pi(d)/\sum_{d'\in\mathcal{T}_L}\pi(d')$. We measure uncertainty
among these diseases as
\begin{equation}
H_L
=
-\sum_{d\in\mathcal{T}_L}
\bar{\pi}(d)\log\bar{\pi}(d).
\end{equation}

Questioning stops when the budget is exhausted or when $H_L<\eta_{\mathrm{stop}}$
after at least $C_{\min}$ phenotypes have been confirmed. The question set may
be reset once to branches not covered by the seeds if the top three diseases
remain unchanged for three rounds without a new confirmation, or if
$H_L>\eta_{\mathrm{reset}}$ after $Q_{\mathrm{reset}}$ questions.

\vspace{2pt} \noindent \textit{{\textbf{Candidate construction and diagnostic synthesis.}}}
\label{sec:retrieval}
\label{sec:synthesis}
Because the ranking induced by $\pi(d\mid\CONF)$ may under-rank diseases
supported by few rare phenotypes, we compute
\begin{equation}
\operatorname{anchor}(d)
=
\sum_{\substack{h\in\CONF\\\pbg(h)\leq\theta_a}}
r(h)P(h\mid d).
\end{equation}
Disease--phenotype associations and frequencies are obtained from
Orphanet~\cite{rath2012orphanet} and HPOA~\cite{gargano2024hpo}; OMIM-only
diseases~\cite{amberger2019omim} without reported phenotype frequencies are
assigned a fixed probability. The synthesis set contains the top $T$ diseases
under $\pi$, up to $A$ previously unseen diseases ranked by
$\operatorname{anchor}(d)$, and the remaining $\pi$-ranked diseases,
retaining the first $R$ unique entries.

After candidate construction, an LLM-generated web query and a
PubCaseFinder~\cite{fujiwara2018pubcasefinder} query are issued from the
confirmed phenotype set. Shared retrieval evidence and each candidate's
Orphanet phenotype profile are then provided to two LLM passes, which produce
a draft of size $N_{\mathrm{draft}}$ and a final ranking of size
$N_{\mathrm{final}}$, which may include diseases outside the synthesis set.

\begin{table*}[t]
\centering
\caption{
Overall diagnosis results on the four RareBench cohorts (\%).
}
\label{tab:main-results}
\setlength{\tabcolsep}{3.2pt}
\resizebox{\textwidth}{!}{%
\begin{tabular}{
l
ccc@{\hspace{7pt}}
ccc@{\hspace{7pt}}
ccc@{\hspace{7pt}}
ccc
}
\toprule
& \multicolumn{3}{c}{MME ($n=40$)}
& \multicolumn{3}{c}{HMS ($n=88$)}
& \multicolumn{3}{c}{LIRICAL ($n=370$)}
& \multicolumn{3}{c}{RAMEDIS ($n=624$)} \\
\cmidrule(lr){2-4}
\cmidrule(lr){5-7}
\cmidrule(lr){8-10}
\cmidrule(lr){11-13}
Method
& R@1 & R@3 & R@5
& R@1 & R@3 & R@5
& R@1 & R@3 & R@5
& R@1 & R@3 & R@5 \\
\midrule

\multicolumn{13}{l}{
\textit{Complete phenotype profile: all recorded HPO terms}
} \\

PhenoBrain\cite{mao2025phenotype}
& 25.0 & 45.0 & 80.0
& 22.4 & 32.9 & 42.4
& 27.1 & 44.7 & 57.7
& 20.7 & 38.3 & 55.8 \\

PubCaseFinder\cite{fujiwara2018pubcasefinder}
& 40.0 & 57.5 & 57.5
& 7.3 & 13.4 & 15.9
& 37.8 & 47.0 & 48.9
& 25.8 & 32.6 & 40.0 \\

GPT-4o\cite{hurst2024gpt}
& 2.5 & 5.0 & 10.0
& 47.7 & 60.2 & 65.9
& 31.1 & 42.2 & 48.9
& 35.5 & 52.2 & 61.1 \\

DeepSeek R1\cite{guo2025deepseek}
& 35.0 & 60.0 & 65.0
& 38.6 & 53.4 & 60.2
& 36.8 & 45.1 & 47.6
& 33.2 & 47.6 & 53.2 \\


DeepRare (GPT-4o)
& 42.5 & 57.5 & 57.7
& 50.0 & 62.5 & 64.8
& 39.5 & 49.2 & 52.4
& 36.9 & 46.4 & 55.4 \\

DeepRare (Qwen-32B)
& 17.5 & 22.5 & 35.0
& 25.0 & 38.6 & 47.7
& 39.5 & 47.6 & 51.9
& 45.0 & 61.2 & 65.4 \\

\midrule

\multicolumn{13}{l}{
\textit{Sparse phenotype profile: three initial HPO terms}
} \\

DeepRare (Qwen-32B)
& 5.0 & 5.0 & 5.0
& 12.5 & 18.2 & 22.7
& \textbf{28.6} & 35.1 & 39.2
& \textbf{33.2} & 45.0 & 48.9 \\

Qwen-32B + web search
& 2.5 & 5.0 & 5.0
& \textbf{18.2} & 20.5 & \textbf{29.5}
& 27.0 & 34.9 & 39.5
& 29.8 & 44.7 & \textbf{52.6} \\

HPOQuest (Qwen-32B)
& \textbf{35.0} & \textbf{42.5} & \textbf{50.0}
& 15.9 & \textbf{23.9} & 25.0
& \textbf{28.6} & \textbf{40.3} & \textbf{45.1}
& 27.6 & \textbf{45.4} & 49.0 \\

\bottomrule
\end{tabular}%
}
\end{table*}

%% file: chapters/experemental_setup.tex
\section{Experimental Setup}
\label{sec:setup-exp}

\paragraph{\textbf{Datasets, baselines, and policies.}}
\label{sec:baselines}
We evaluate on the four RareBench~\cite{chen2024rarebench} cohorts:
MME ($n=40$), HMS ($n=88$), LIRICAL ($n=370$), and RAMEDIS ($n=624$).
All sparse-input methods receive the same three seed phenotypes. For patients with more than three recorded phenotypes, we select the two phenotypes occurring in the fewest Orphanet disease profiles and sample one additional phenotype uniformly from the remaining terms using a deterministic patient-specific random seed; patients with at most three recorded phenotypes use all available terms. DeepRare~\cite{zhao2026agentic} receives either these seeds or the complete
phenotype profile. Qwen-32B + web receives only the seeds and web
access, without phenotype acquisition or candidate retrieval. All controlled
systems use the same diagnoser and judge, with DeepRare's private similar-case
retrieval disabled. DeepRare with the complete phenotype profile and published
full-profile results serve only as references.

We compare CAR with three acquisition policies under the same experimental
setting. LLM-guided selection inspired by
MediQ~\cite{li2024mediq} prompts an expert LLM to select from the current HPO
candidate set or abstain. One-step expected information gain
(EIG)~\cite{lindley1956measure} ranks candidate terms by their expected entropy
reduction over the current top-$M$ diseases using $P(h\mid d)$; only the
ranking policy changes, while responses, including \textsc{yes-branch}, follow
Section~\ref{sec:frontier-update}. CAR--EIG combines the CAR and EIG rankings
using reciprocal rank fusion (RRF)~\cite{cormack2009reciprocal} with
$\kappa=60$.

\vspace{2pt} 
\noindent\textit{\textbf{Implementation details.}}
\label{sec:hyper}
We use $K=3$ seed phenotypes, question budget $B=20$, per-round limit $b=8$,
and $M=50$ diseases for question scoring. Stopping uses
$(L,\eta_{\mathrm{stop}},C_{\min},\eta_{\mathrm{reset}},Q_{\mathrm{reset}})
=(10,0.8,4,1.2,10)$. CAR uses
$(\theta_s,\gamma,\lambda,\beta,\phi,\tau,\theta_f,\theta_r,\theta_a)
=(0.02,0.7,4,2,0.002,0.02,0.55,0.02,0.01)$.
Candidate construction uses $(T,A,R)=(8,4,12)$.
Missing HPOA frequencies are assigned $P(h\mid d)=0.5$. Anchor retrieval covers 4,283 Orphanet and 9,120 HPOA diseases.
The synthesis passes use
$(N_{\mathrm{draft}},N_{\mathrm{final}})=(12,15)$.
All controlled systems use 4-bit Qwen2.5-32B-Instruct on one GPU, SerpAPI for
web search, and GPT-4o-mini with DeepRare's judging prompt. Runtime per
patient is 370\,s (CAR), 384\,s (CAR--EIG), and 496\,s/611\,s (DeepRare with
three/all phenotypes). We report Recall@$k$ ($k\in\{1,3,5\}$), accepting
predictions verified by either the LLM judge or a MONDO \cite{vasilevsky2020mondo} synonym.

%% file: chapters/results_dis.tex
\begin{table*}[t]
\centering
\caption{
Acquisition-policy ablation on the four RareBench cohorts (\%).
}
\label{tab:policy-ablation}
\setlength{\tabcolsep}{3.2pt}
\resizebox{\textwidth}{!}{%
\begin{tabular}{
l
ccc@{\hspace{7pt}}
ccc@{\hspace{7pt}}
ccc@{\hspace{7pt}}
ccc
}
\toprule
& \multicolumn{3}{c}{MME ($n=40$)}
& \multicolumn{3}{c}{HMS ($n=88$)}
& \multicolumn{3}{c}{LIRICAL ($n=370$)}
& \multicolumn{3}{c}{RAMEDIS ($n=624$)} \\
\cmidrule(lr){2-4}
\cmidrule(lr){5-7}
\cmidrule(lr){8-10}
\cmidrule(lr){11-13}
Configuration
& R@1 & R@3 & R@5
& R@1 & R@3 & R@5
& R@1 & R@3 & R@5
& R@1 & R@3 & R@5 \\
\midrule

\multicolumn{13}{l}{\textit{No acquisition, Qwen-32B, 3 HPOs}} \\

DeepRare
& 5.0 & 5.0 & 5.0
& 12.5 & 18.2 & 22.7
& 28.6 & 35.1 & 39.2
& 33.2 & 45.0 & 48.9 \\

LLM + web search
& 2.5 & 5.0 & 5.0
& 18.2 & 20.5 & 29.5
& 27.0 & 34.9 & 39.5
& 29.8 & 44.7 & 52.6 \\

\midrule
\multicolumn{13}{l}{\textit{HPOQuest acquisition policies}} \\

LLM-guided selection
& 27.5 & 40.0 & 42.5
& \textbf{20.5} & \textbf{28.4} & \textbf{30.7}
& 30.0 & 39.5 & 43.8
& 26.4 & 43.4 & 47.9 \\

EIG
& 22.5 & 37.5 & 37.5
& 18.2 & \textbf{28.4} & \textbf{30.7}
& \textbf{31.4} & 39.5 & 43.0
& 27.9 & 43.6 & 47.4 \\

CAR
& \textbf{35.0} & 42.5 & \textbf{50.0}
& 15.9 & 23.9 & 25.0
& 28.6 & \textbf{40.3} & \textbf{45.1}
& 27.6 & \textbf{45.4} & 49.0 \\

RRF
& \textbf{35.0} & \textbf{47.5} & 47.5
& \textbf{20.5} & 26.1 & 28.4
& 26.5 & 37.8 & 43.0
& \textbf{29.0} & 42.8 & \textbf{49.7} \\

\bottomrule
\end{tabular}%
}
\end{table*}

\section{Results and Discussion}
\noindent \textit{\textbf{Overall performance.}}
Active phenotype acquisition provides its clearest benefit on MME
(Table~\ref{tab:main-results}), where HPOQuest with CAR improves
Recall@1/3/5 from $5.0/5.0/5.0\%$ to $35.0/42.5/50.0\%$, exceeding the
Qwen-32B full-profile DeepRare reference despite starting from only three seed
phenotypes. This indicates that acquiring a small number of informative
phenotypes can be more effective than relying on the complete recorded profile
when the initial evidence is insufficient for diagnosis. On LIRICAL, CAR
improves Recall@3 and Recall@5 while matching DeepRare-3 at Recall@1. Gains
are smaller on HMS and RAMEDIS, where the strong sparse-input baselines,
particularly Qwen-32B + web, suggest that the language model can already infer
a plausible disease region from the initial phenotypes. In these cohorts,
active acquisition primarily refines the differential diagnosis, yielding more
consistent improvements at Recall@3 and Recall@5 than at Recall@1.

\vspace{2pt} \noindent \textit{\textbf{Acquisition policies.}}
No single acquisition policy dominates across cohorts
(Table~\ref{tab:policy-ablation}). CAR performs best at deeper cutoffs on MME
and LIRICAL, RRF achieves the highest Recall@3 on MME and Recall@1/5 on
RAMEDIS, while EIG and LLM-guided selection perform best on HMS. These trends
reflect the complementary objectives of the policies: CAR prioritizes phenotype
specificity within the current candidate set, EIG explicitly reduces posterior
uncertainty, and LLM-guided selection leverages the language model's prior
biomedical knowledge. The results therefore suggest that the optimal
acquisition strategy depends on the diagnostic setting rather than one policy
being uniformly superior.

\vspace{2pt} \noindent \textit{\textbf{Seed recoverability.}}
Table~\ref{tab:seed-recoverability} stratifies patients according to seed
recoverability, defined as whether the reference disease appears among the
top-15 Orphanet candidates from the three seed phenotypes alone. Acquisition
is most effective for seed-recoverable cases, raising Recall@1 from $0.0\%$ to
$77.8\%$ on MME and from $39.2\%$ to $47.3\%$ on LIRICAL, with larger gains at
deeper cutoffs. For non-recoverable cases, improvements remain limited across
cohorts, suggesting that active questioning is most effective when the initial
seed phenotypes already provide sufficient signal to identify a plausible
diagnostic region. All reported $0.0\%$ entries are observed
results, indicating that no patient in the corresponding subgroup was ranked.
Taken together, the results reveal two operating regimes.
When the initial phenotypes provide only a weak signal for the language model,
as on MME, active phenotype acquisition substantially improves diagnostic
accuracy. In contrast, when the language model can already infer a plausible
disease region from the seed phenotypes, as suggested by the strong Qwen-32B +
web baselines on HMS, LIRICAL, and RAMEDIS, acquisition primarily refines the
ordering of the differential diagnosis rather than changing the top-ranked
prediction.
\begin{table*}[t]
\centering
\caption{
Results stratified by seed recoverability (\%).
\textsc{R} denotes seed-recoverable cases and \textsc{NR} non-recoverable
cases. All configurations use Qwen-32B.
}
\label{tab:seed-recoverability}
\setlength{\tabcolsep}{3.5pt}
\resizebox{\textwidth}{!}{%
\begin{tabular}{
lcr
ccc@{\hspace{7pt}}
ccc@{\hspace{7pt}}
ccc@{\hspace{7pt}}
ccc
}
\toprule
&&&
\multicolumn{3}{c}{DeepRare-All}
& \multicolumn{3}{c}{DeepRare-3}
& \multicolumn{3}{c}{Qwen-32B + web}
& \multicolumn{3}{c}{HPOQuest (CAR)} \\
\cmidrule(lr){4-6}
\cmidrule(lr){7-9}
\cmidrule(lr){10-12}
\cmidrule(lr){13-15}
Cohort & Group & $n$
& R@1 & R@3 & R@5
& R@1 & R@3 & R@5
& R@1 & R@3 & R@5
& R@1 & R@3 & R@5 \\
\midrule

MME
& \textsc{R}
& 9
& 0.0 & 0.0 & 0.0
& 0.0 & 0.0 & 0.0
& 0.0 & 0.0 & 0.0
& 77.8 & 77.8 & 88.9 \\

& \textsc{NR}
& 31
& 22.6 & 29.0 & 45.2
& 6.5 & 6.5 & 6.5
& 3.2 & 6.5 & 6.5
& 22.6 & 32.3 & 38.7 \\

\addlinespace

HMS
& \textsc{R}
& 7
& 28.6 & 42.9 & 57.1
& 14.3 & 28.6 & 28.6
& 0.0 & 14.3 & 28.6
& 0.0 & 57.1 & 57.1 \\

& \textsc{NR}
& 81
& 24.7 & 38.3 & 46.9
& 12.3 & 17.3 & 22.2
& 19.8 & 21.0 & 29.6
& 17.3 & 21.0 & 22.2 \\

\addlinespace

LIRICAL
& \textsc{R}
& 74
& 54.1 & 62.2 & 68.9
& 39.2 & 47.3 & 56.8
& 43.2 & 58.1 & 66.2
& 47.3 & 73.0 & 81.1 \\

& \textsc{NR}
& 296
& 35.8 & 43.9 & 47.6
& 26.0 & 32.1 & 34.8
& 23.0 & 29.1 & 32.8
& 24.0 & 32.1 & 36.1 \\

\addlinespace

RAMEDIS
& \textsc{R}
& 114
& 52.6 & 76.3 & 80.7
& 46.5 & 78.1 & 78.1
& 44.7 & 68.4 & 86.0
& 40.4 & 87.7 & 91.2 \\

& \textsc{NR}
& 510
& 43.3 & 57.8 & 62.0
& 30.2 & 37.6 & 42.4
& 26.5 & 39.4 & 45.1
& 24.7 & 35.9 & 39.6 \\

\bottomrule
\end{tabular}%
}
\end{table*}

%% file: chapters/conclusion.tex
\section{Conclusion}
We introduced HPOQuest, a training-free framework for active HPO acquisition
that formulates rare-disease diagnosis as sequential selection of
ontology-constrained phenotype questions. Across four RareBench cohorts,
HPOQuest substantially improves diagnosis when sparse initial phenotypes
provide limited diagnostic evidence, while remaining competitive when the
initial phenotype signal is already informative. Our analyses further show that
the effectiveness of active acquisition depends on both the informativeness of
the initial phenotypes and the strength of the language model's prior
knowledge, suggesting that structured questioning primarily complements
existing LLM capabilities by refining the differential diagnosis. Future work should investigate robustness across different initial
phenotype settings and language models.